\documentclass[10pt,twocolumn,letterpaper]{article}

\usepackage[pagenumbers]{cvpr} 

\definecolor{cvprblue}{rgb}{0.21,0.49,0.74}
\usepackage[pagebackref,breaklinks,colorlinks,allcolors=cvprblue]{hyperref}
\usepackage[ruled, linesnumbered]{algorithm2e}
\usepackage{booktabs}
\usepackage{colortbl}
\usepackage{booktabs}
\usepackage{multirow}
\usepackage{bm} 
\usepackage{siunitx} 
\makeatletter

\newcommand{\Rmnum}[1]{\expandafter\@slowromancap\romannumeral #1@}
\usepackage{pifont}
\usepackage{multirow}

\definecolor{lightred}{rgb}{1.0, 0.78, 0.81}
\definecolor{lightpurple}{rgb}{0.88, 0.77, 1}

\def\paperID{6836} 
\def\confName{CVPR}
\def\confYear{2026}

\title{VideoFusion: A Spatio-Temporal Collaborative Network \\ for Multi-modal Video Fusion}

\author{
	\begin{tabular}{c}
		Linfeng Tang$^{1}$,
		Yeda Wang$^{2}$,
		Meiqi Gong$^{2}$,
		Zizhuo Li$^{2}$,
		Yuxin Deng$^{2}$\\
		Xunpeng Yi$^{2}$,
		Chunyu Li$^{2}$,
		Hao Zhang$^{2}$,
		Han Xu$^{3}$,
		Jiayi Ma$^{1}$\thanks{Corresponding author.}\\		
		$^{1}$School of Robotics, Wuhan University, Wuhan, China\\
		$^{2}$Electronic Information School, Wuhan University, Wuhan, China\\
		$^{3}$Southeast University, Nanjing, China\\	
		{\ttfamily\small \{linfeng0419,licy0089,zhpersonalbox,jyma2010\}@gmail.com}\\
		{\ttfamily\small \{wangyeda,meiqigong,zizhuo\_li,dyx\_acuo,yixunpeng\}@whu.edu.cn},
		{\ttfamily\small xu\_han@seu.edu.cn}		
	\end{tabular}
}
\begin{document}
\maketitle
\begin{abstract}
	Compared to images, videos better reflect real-world acquisition and possess valuable temporal cues. However, existing multi-sensor fusion research predominantly integrates complementary context from multiple images rather than videos due to the scarcity of large-scale multi-sensor video datasets, limiting research in video fusion and the inherent difficulty of jointly modeling spatial and temporal dependencies in a unified framework. To this end, we construct \textbf{M3SVD}, a benchmark dataset with $220$ temporally synchronized and spatially registered infrared-visible videos comprising $153,797$ frames, bridging the data gap. Secondly, we propose \textbf{VideoFusion}, a multi-modal video fusion model that exploits cross-modal complementarity and temporal dynamics to generate spatio-temporally coherent videos from multi-modal inputs. Specifically, 1) a differential reinforcement module is developed for cross-modal information interaction and enhancement, 2) a complete modality-guided fusion strategy is employed to adaptively integrate multi-modal features, and 3) a bi-temporal co-attention mechanism is devised to dynamically aggregate forward-backward temporal contexts to reinforce cross-frame feature representations. Experiments reveal that \textbf{VideoFusion} outperforms existing image-oriented fusion paradigms in sequences, effectively mitigating temporal inconsistency and interference. Project and M3SVD:~\url{https://github.com/Linfeng-Tang/VideoFusion}.
\end{abstract}


\section{Introduction} \label{sec:intro}

Single-type sensors capture information from a limited perspective, making it challenging to comprehensively characterize imaging scenarios~\cite{Zhang2021survey}. For instance, visible sensors rely on reflected light to capture detailed textures but are vulnerable to environmental factors. In contrast, infrared sensors leverage thermal radiation to highlight salient targets effectively, yet they lack fine-grained texture representation. Consequently, multi-sensor fusion, which integrates complementary information from diverse sensors to overcome the limitations of single-type sensors, has garnered considerable attention~\cite{Karim2023Survey}. Among these, infrared and visible sensor fusion has emerged as a prominent research focus, demonstrating great potential in military detection~\cite{Muller2009Military}, security surveillance~\cite{Zhang2018Surveillance}, assisted driving~\cite{Bao2023Driving}.

\begin{figure}[t]
	\centering
	\includegraphics[width=0.97\linewidth]{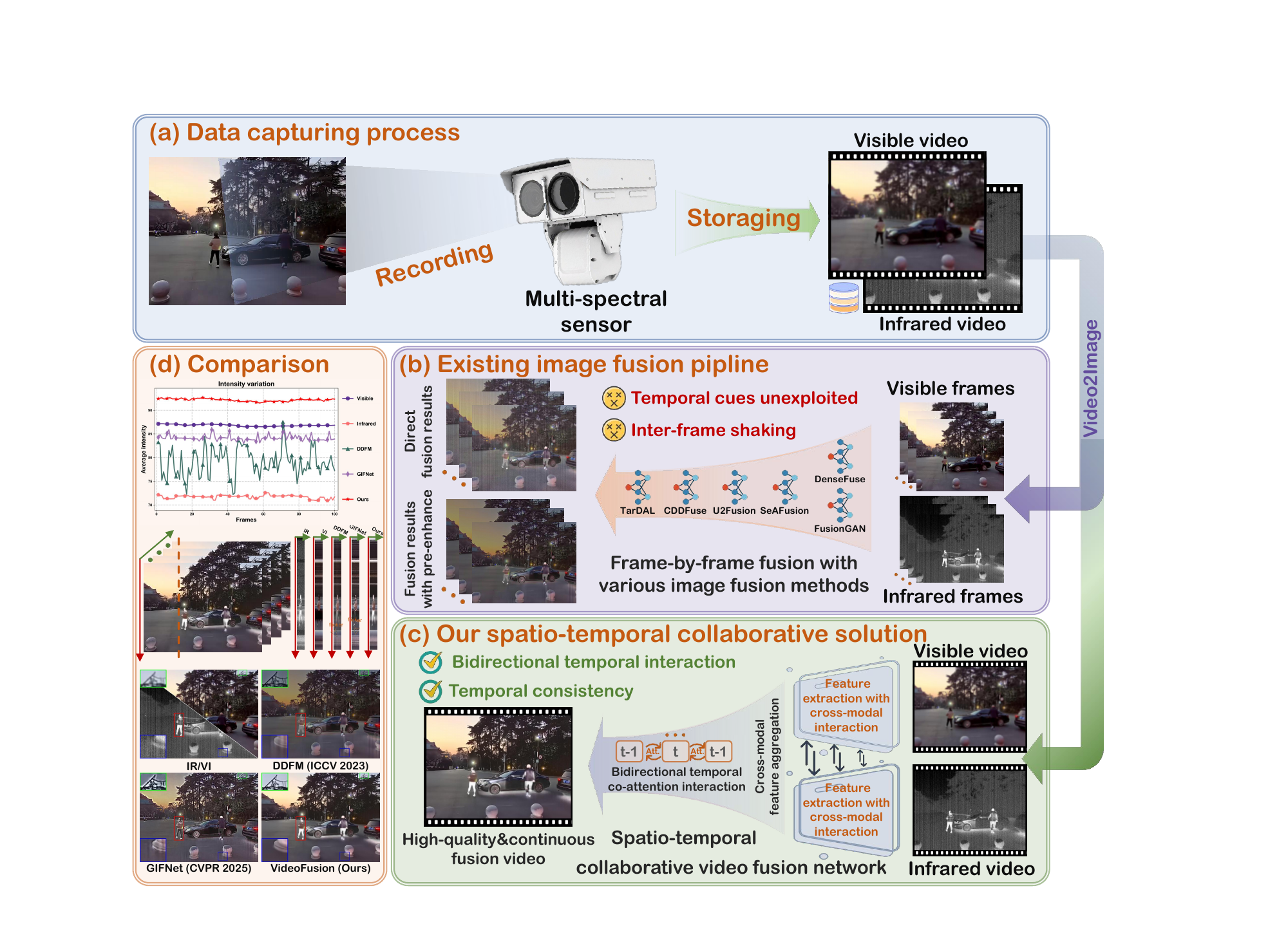}	
	\caption{Image-oriented fusion vs. video fusion.}
	\label{fig:example}
\end{figure}

Earlier multi-sensor fusion studies primarily focused on image fusion, introducing various sophisticated networks, such as auto-encoders~\cite{Li2018DenseFuse}, convolutional neural networks~\cite{Liu2024TIMFusion}, generative adversarial networks~\cite{Ma2019FusionGAN}, Transformers~\cite{Ma2022SwinFusion}, and diffusion models~\cite{Zhao2023DDFM}, in pursuit of superior visual perception. Moreover, some semantic-driven methods~\cite{Tang2022SeAFusion}, considering semantic demands of downstream tasks, are proposed to reinforce the bond between image fusion and high-level vision tasks\cite{Liu2023FMB, Zhao2023MetaFusion}. Besides, some joint registration and fusion proposals~\cite{Xu2023MURF}, and degradation-robust solutions~\cite{Tang2024DRMF} are also designed to cope with interference factors encountered in the imaging process. Particularly, degradation-robust methods leverage cross-modal complementary context to suppress degradation and directly generate high-quality fusion results~\cite{Zhang2024DDBF}, expanding the application scope of image fusion.

Although these methods achieve satisfactory performance, they neglect a crucial aspect that in practical applications, multi-modal sensors typically capture continuous video sequences rather than independent static frames, as illustrated in Fig.~\ref{fig:example}~(a). However, existing multi-sensor fusion methods are tailored for static images, primarily due to \emph{the lack of large-scale multi-modal video datasets}. Consequently, they exploit cross-modal complementary features while overlooking inherent temporal dependencies in videos. It is worth noting that naively extending image fusion to frame-wise video fusion (Fig.~\ref{fig:example}~(b)) ignores inter-frame complementarity and introduces temporal incoherence, leading to inter-frame flickering artifacts, as shown in Fig.~\ref{fig:example}~(d). Furthermore, recent advances in video restoration have demonstrated that temporal cues play a crucial role in mitigating visual degradations such as deblurring~\cite{Rao2024VD-Diff}, deraining~\cite{Lin2024NightRain}, and low-light enhancement~\cite{Lv2023Unsupervised}.

In this work, we first construct \textbf{M3SVD}, a multi-modal multi-scene video dataset with 220 temporally synchronized and spatially aligned infrared-visible videos totaling $153,797$ frames. Moreover, we propose a spatio-temporal collaborative video fusion network (\textbf{VideoFusion}), as presented in Fig.~\ref{fig:example}(c), to exploit temporal cues and cross-modal complementarity for robust representation. On the one hand, existing studies indicate that different modalities are typically complementary but also contain redundant information~\cite{Yang2024STCNet}. Thus, we design a cross-modal differential reinforcement module that exploits the complementary yet non-redundant information across modalities to achieve cross-modal information interaction and enhancement. Additionally, we develop a complete modality-guided fusion module to integrate enhanced features, which employs the sum of comprehensive infrared and visible features as a query to aggregate cross-modal complementarity. On the other hand, recognizing that cross-frame features offer dynamic cues to enrich scene representation, we devise a bidirectional temporal co-attention mechanism to integrate informative features from neighboring frames in both forward and backward directions into the current frame. By combining forward and backward attention, it effectively leverages temporal cues. To this end, the proposed method leverages complementary information across modalities and exploits temporal cues across frames to produce high-quality fused videos from degraded inputs, as illustrated in Fig.~\ref{fig:example}(d). In summary, our main contributions are as follows:
\begin{itemize}	
	\item[-] A multi-modal multi-scene video dataset~(M3SVD), comprising $220$ temporally synchronized and spatially registered infrared-visible videos with $153,797$ frames, is constructed as the large-scale benchmark for video fusion, restoration, registration, and related areas.
	\item[-] We propose a spatio-temporal collaborative network for infrared and visible video fusion, pioneering the joint modeling of cross-modal complementarity and across-frame temporal cues, advancing video fusion tasks.
	\item[-] We devise a bi-temporal co-attention mechanism with a variational consistency loss to exploit forward and backward temporal cues, coupled with a differential reinforcement module to harness cross-modal complementary context for information restoration and integration.
\end{itemize}

\section{Related Work} \label{sec:related}
\subsection{Image Fusion}
At an early stage, image fusion methods primarily focus on aggregating cross-modal complementary information to enhance visual perception. Elaborate network architectures, including CNN~\cite{Xu2022U2Fusion, Zhu2024TC-MoA}, AE~\cite{Li2018DenseFuse, Li2023LRRNet}, GAN~\cite{Liu2022TarDAL, Zhang2024DDBF}, Transformer~\cite{Ma2022SwinFusion, Zhao2023CDDFuse}, and diffusion models~\cite{Zhao2023DDFM, Tang2025Mask-DiFuser}, along with loss functions, such as intensity~\cite{Ma2021STDFusionNet}, gradient~\cite{Zhang2020PMGI}, SSIM~\cite{Ma2022SwinFusion} and perceptual~\cite{Zhang2020IFCNN} losses, are employed to preserve source-consistent meaningful information, effectively enhancing human visual perception.
In particular, considering semantic requirements of downstream tasks, researchers proposed semantic-driven algorithms that leverage semantic segmentation~\cite{Tang2022SeAFusion, Liu2023FMB, Zhang2024MRFS} or object detection~\cite{Zhao2023MetaFusion, Liu2022TarDAL} to enhance semantic preservation in fusion networks. Moreover, integrated registration and fusion frameworks~\cite{Tang2022SuperFusion, Xu2023MURF, Li2025MulFS-CAP} are introduced to simultaneously achieve alignment and fusion, mitigating parallax and distortion in real-world imaging and reducing artifacts in fusion results. Furthermore, degradation-robust paradigms~\cite{Yi2024Text-IF, Tang2024DRMF, Zhang2025OmniFuse, Tang2025ControlFusion} are proposed to counteract degradation interference in the imaging process. However, these algorithms only leverage limited information available in static images to enrich scene representation, thereby failing to fully harness the potential of temporal cues inherent in video sequences.

\subsection{Video Fusion}
A few studies have recently focused on video fusion, such as RCVS~\cite{Xie2024RCVS}, saPIDFuse~\cite{Tang2024saPIDFuse}, and DFTP~\cite{Guo2024DFTP}. Specifically, Xie~\emph{et al.} developed a unified registration and fusion framework for video streams~\cite{Xie2024RCVS}. However, these methods either adopt frame-by-frame strategies or rely on handcrafted features for temporal modeling, thus limiting the exploitation of temporal cues and harming temporal consistency. Our concurrent works, including TemCoCo~\cite{Gong2025TemCoCo} and UniVF~\cite{Zhao2025UniVF}, attempt to directly aggregate video sequences to alleviate inter-frame jitter. However, they depend on DCN or optical flow for inter-frame information compensation. DCN estimates frame-to-frame offsets in an unsupervised manner, which may lead to instability, while optical flow networks, typically trained on visible images, struggle to generalize across multi-modal datasets. This work explores the potential of \textit{attention mechanisms} to adaptively aggregate cross-modal and temporal complementary information, providing a more comprehensive and robust scene representation.


\begin{table}[t]
	\centering
	\setlength{\tabcolsep}{2pt}
	\caption{Comparison of aligned multi-modal datasets.} \label{tab:dataset}
	\resizebox{0.48\textwidth}{!}{
		\begin{tabular}{lcccccccc}
			\toprule
			\multirow{2}{*}{\textbf{Datasets}} & \multirow{2}{*}{\textbf{Temporal}} & \multirow{2}{*}{\textbf{Video nums}} & \multirow{2}{*}{\textbf{Image pairs}} & \multirow{2}{*}{\textbf{Resolution}} & \multicolumn{4}{c}{\textbf{Challenging scenarios}} \\
			\cmidrule(lr){6-9}
			&  &  &  &  &\textbf{Low-light}  & \textbf{Over-exp.} & \textbf{Disguise} & \textbf{Occlusion} \\
			\midrule
			\textbf{RoadScene}  & No  & --    & 221    & 768$\times$576  & \ding{51} & \ding{51} & \ding{55} & \ding{55} \\
			\textbf{MSRS}        & No  & --    & 1,444  & 640$\times$480  & \ding{51} & \ding{51} & \ding{55} & \ding{55} \\
			\textbf{M}$^3$\textbf{FD}     & No  & --    & 4,200  & 1024$\times$768 & \ding{51} & \ding{51} & \ding{55} & \ding{51} \\
			\textbf{FMB}         & No  & --    & 1,500  & 800$\times$600  & \ding{51} & \ding{51} & \ding{55} & \ding{51} \\
			\textbf{LLVIP}       & No  & --    & 15,488 & \textbf{1280$\times$1024} & \ding{51} & \ding{51} & \ding{55} & \ding{55} \\
			\textbf{TNO}         & Yes & 3     & 114    & 768$\times$576  & \ding{51} & \ding{55} & \ding{55} & \ding{51} \\
			\textbf{INO}    & Yes & 15    & 12,695 & 328$\times$254  & \ding{55} & \ding{51} & \ding{55} & \ding{55} \\
			\textbf{HDO}        & Yes & 24    & 7,500  & 640$\times$480  & \ding{51} & \ding{51} & \ding{55} & \ding{55} \\
			\rowcolor{lightpurple!30}  
			\textbf{M3SVD}       & Yes & \textbf{220}   & \textbf{153,797}& 640$\times$480  & \ding{51} & \ding{51} & \ding{51} & \ding{51} \\
			\bottomrule
		\end{tabular}
	}
\end{table}

\begin{figure}[t]
	\centering
	\includegraphics[width=1\linewidth]{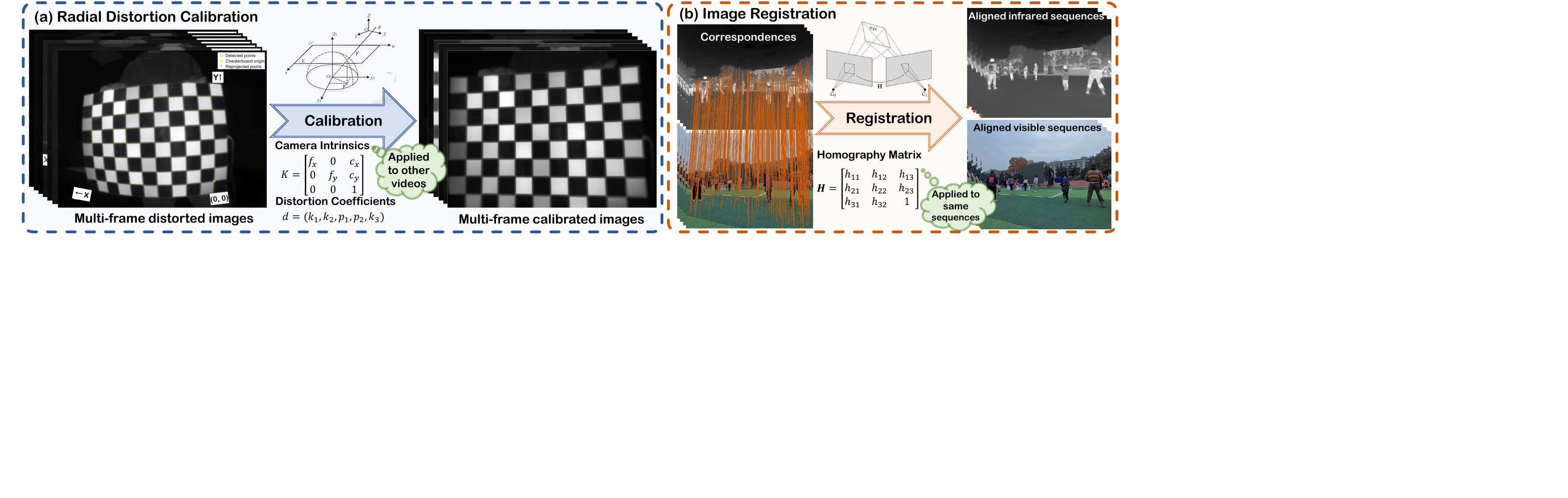}	
	\vspace{-0.15in}
	\caption{Schematic of image calibration and registration.}
	\label{fig:Workflow}
\end{figure}

\section{Multi-Modal Multi-Scene Video Dataset}
As shown in Tab.~\ref{tab:dataset}, although multiple multi-modal datasets have been developed for image fusion tasks, most existing datasets (\emph{e.g.}, RoadScene~\cite{Xu2022U2Fusion}, MSRS~\cite{Tang2022PIAFusion}, M$^3$FD~\cite{Liu2022TarDAL}, and FMB~\cite{Liu2023SegMiF}) focus on static image pairs, lacking temporal coherence and limiting their applicability to dynamic scenarios. Only a small subset (\emph{e.g.}, TNO~\cite{Toet2017TNO}, INO, and  HDO~\cite{Xie2024RCVS}) incorporate temporal information, yet these video datasets remain scarce and critically underdeveloped. For instance, TNO has limited data, INO suffers from low resolution, and HDO exhibits poor imaging quality, with all covering only a narrow range of scenarios. Additional dataset comparisons are included in the \textbf{Suppl. Material}.

Therefore, we construct a multi-modal multi-scene video dataset~(M3SVD) with a synchronized dual-spectral imaging system\footnote{https://www.magnity.com.cn/product/id/76.html}, to support standardized training and evaluation of multi-modal video fusion. The acquisition setup comprises: 1) an uncooled infrared sensor ($7.5\sim 14 \mu$m spectral response, $640 \times 480$ resolution, @30FPS), and 2) a visible CMOS sensor ($1920 \times 1080$ resolution, @30FPS). As shown in Fig.~\ref{fig:Workflow}, we first employ the \textit{Matlab Camera Calibrator} with the checkerboard calibration plate to correct the radial distortion in both sensors. Second, due to the non-coincident optical centers and scale variations between two sensors, image registration is necessary. Given the $3.5$cm inter-optical-axis distance between the sensors, the transformation between infrared and visible frames can be effectively approximated by a homography matrix~$\textbf{H}$. To enhance the stability of the registered videos, we estimate a single $\textbf{H}$ for one video pair. We uniformly sample image pairs from multi-modal videos to establish precise correspondences using ReDFeat~\cite{Deng2022ReDFeat}, XCP-Match~\cite{Yang2025XCP-Match}, and MINIMA~\cite{Jiang2024MINIMA}, followed by MAGSAC++~\cite{Barath2020MAGSAC++} to estimate $\textbf{H}$. Finally, we register each infrared frame to the corresponding downsampled visible frame using $\textbf{H}$, resulting in infrared and visible videos with a resolution of $640\times480$ at $30$ FPS.

\begin{figure}[t]
	\centering
	\includegraphics[width=0.86\linewidth]{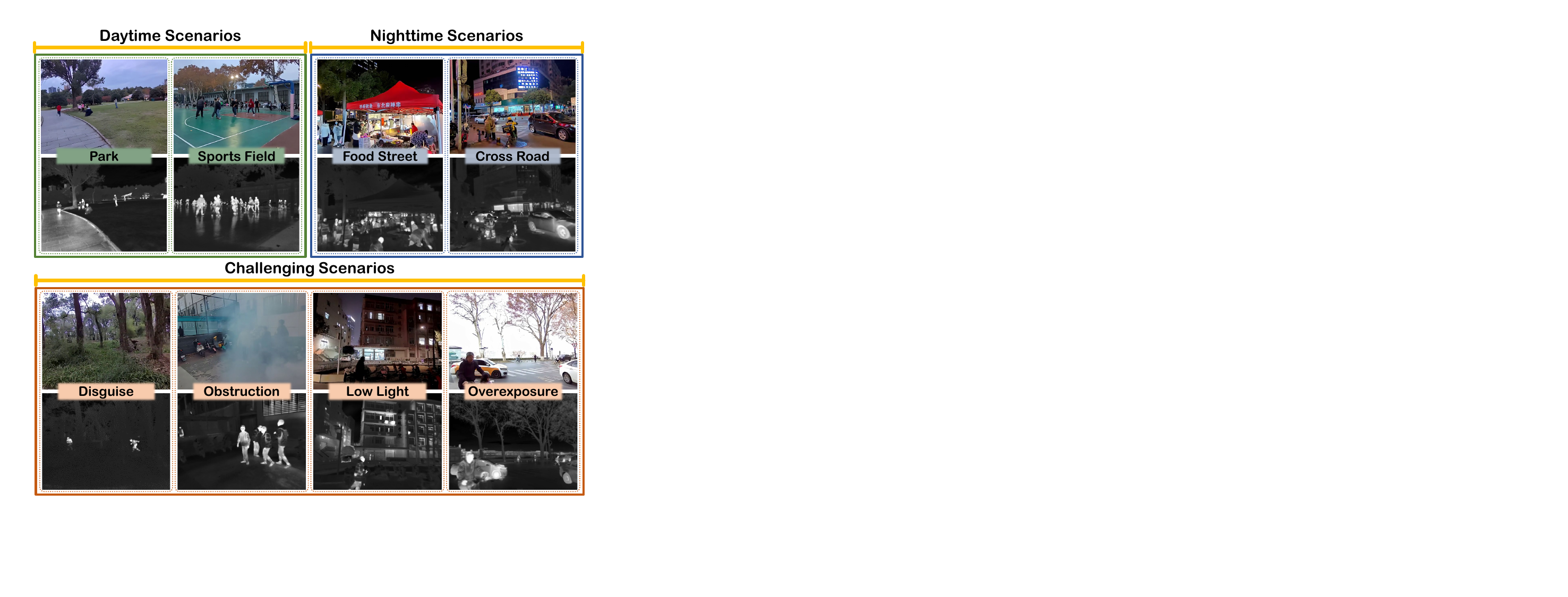}
	\vspace{-0.08in}
	\caption{Visualization of various scenarios in M3SVD dataset.}
	\label{fig:Dataset}
\end{figure}

As shown in Fig.~\ref{fig:Dataset}, M3SVD encompasses daytime, nighttime, and challenging scenarios, covering diverse locations such as parks, lakes, sports fields, food streets, and crossroads. The challenging scenes involve typical multi-modal application surroundings such as disguise, occlusion, low light, and overexposure. In total, we collect $220$ time-synchronized and spatially aligned multi-modal videos with $153,797$ frames, spanning $100$ distinct scenes.

\begin{figure*}[t]
	\centering
	\includegraphics[width=0.88\linewidth]{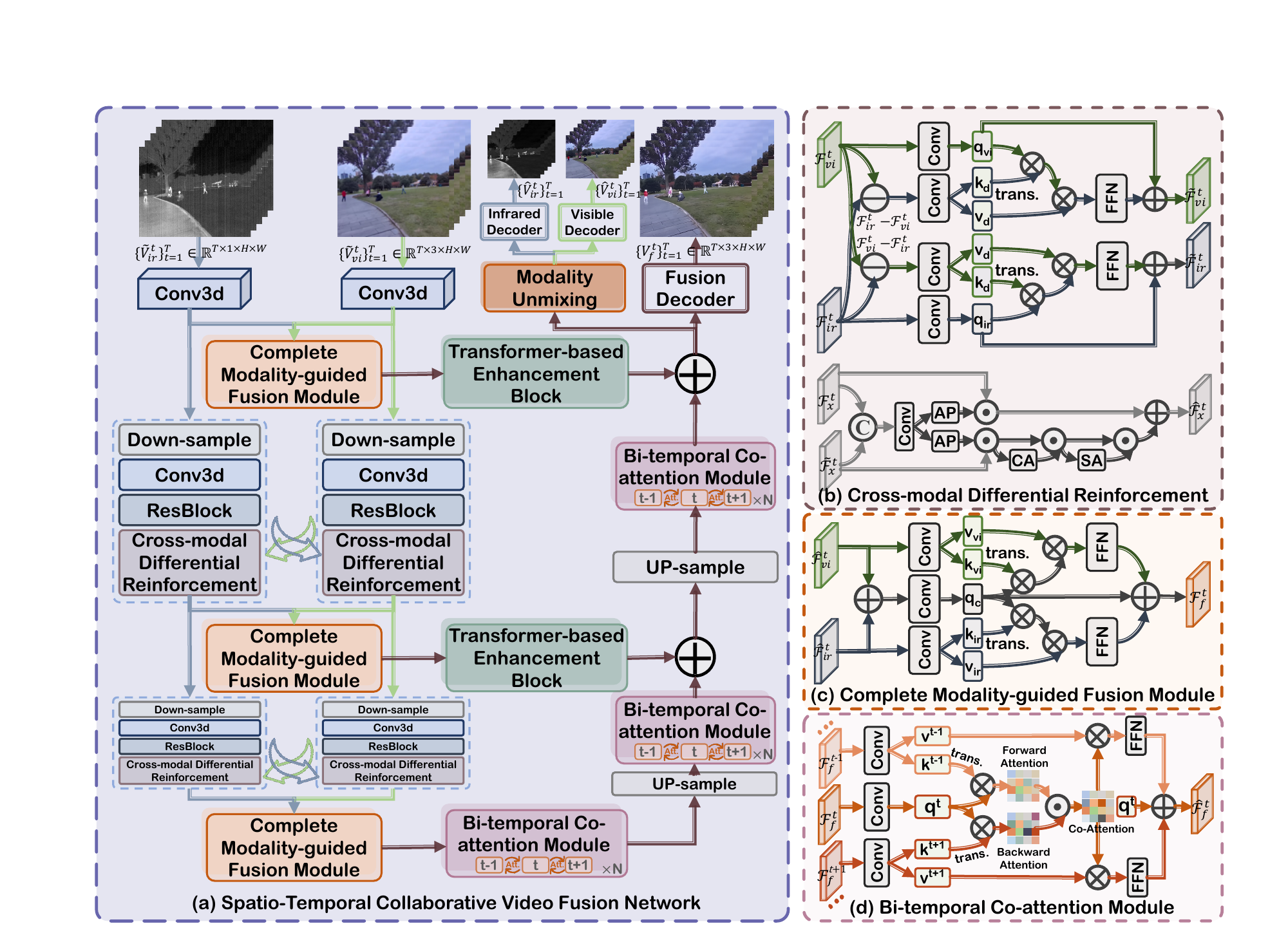}	
	\caption{The overall framework of our spatio-temporal collaborative video fusion network.}
	\label{fig:Framework}
\end{figure*}

\section{Methodology}
\subsection{Overview}
Our workflow is illustrated in Fig.~\ref{fig:Framework}~(a). Given low-quality source videos $\{\tilde{V}_{ir}^i\}_{i=1}^T$ ($\tilde{V}_{ir}^i \in \mathbb{R}^{1 \times H \times W}$) and $\{\tilde{V}_{vi}^i\}_{i=1}^T$ ($\tilde{V}_{vi}^i \in \mathbb{R}^{3 \times H \times W}$), we propose a spatio-temporal collaborative network to generate a high-quality fused video $\{V_f^i\}_{i=1}^T$ ($V_f^i \in \mathbb{R}^{3 \times H \times W}$), along with degradation-free infrared and visible videos $\{\hat{V}_{ir}^i\}_{i=1}^T$ and $\{\hat{V}_{vi}^i\}_{i=1}^T$. In the encoding stage, we first apply 3D convolution (Conv3D) to extract shallow temporal features $\mathcal{F}_{x}^{1} \in \mathbb{R}^{T \times C_1 \times H \times W}$, where $x \in \{ir, vi\}$. Then, a sequential architecture composed of down-sampling layers, Conv3D, ResBlock, and a cross-modal differential reinforcement module (CmDRM) is utilized to extract multi-scale temporal features $\mathcal{F}_{x}^{n} \in \mathbb{R}^{T \times C_n \times H_n \times W_n}$ for $n \in \{2, 3\}$. In particular, CmDRM enhances unimodal representations by integrating cross-modal differential information. Subsequently, complete modality-guided fusion modules are employed at multiple scales ($n \in \{1, 2, 3\}$) to aggregate cross-modal context and generate informative fusion features $\mathcal{F}_f^n$. To further enrich the feature representation, transformer-based enhancement blocks are introduced, followed by a bi-temporal co-attention module (BiCAM) in the decoding stage. BiCAM establishes dynamic temporal dependencies that go beyond Conv3D operations. Finally, a fusion decoder reconstructs the high-quality fusion video $\{V_f^i\}_{i=1}^T$, and a modality unmixing module with dedicated infrared and visible decoders yields $\{\hat{V}_{ir}^i\}_{i=1}^T$ and $\{\hat{V}_{vi}^i\}_{i=1}^T$. This design promotes degradation-invariant fusion while preserving cross-modal information.

\subsection{Network Architectures}

\subsubsection{Cross-modal Differential Reinforcement Module} Given that features from various modalities introduce both complementary contexts and redundant information~\cite{Tang2022PIAFusion, Yang2024STCNet}, the cross-modal differential reinforcement module~(CmDRM) first employs a cross-modal attention mechanism to aggregate differential information from complementary modalities, as illustrated in Fig.~\ref{fig:Framework}~(b). For conciseness, we omit scale indices~($n$), hence $\mathcal{F}_{x}^t\in \mathbb{R}^{C\times H \times W} (x \in \{ir, vi\})$ denote infrared or visible features of $t$-th frame. For instance, designating the visible modality as primary and the infrared as auxiliary, the complementary differential features for the primary modality are computed as $\mathcal{F}_{d}^t = \mathcal{F}_{ir}^t - \mathcal{F}_{vi}^t$, which captures unique information in the auxiliary modality. A $1 \times 1$ convolution is then applied to project $\mathcal{F}_{d}^t$ into keys~($k_d$) and values~$v_d$, while mapping $\mathcal{F}_{vi}^t$ as queries~($q_{vi}$). The differentially reinforced features can be formulated as:
\begin{equation}
	\tilde{\mathcal{F}}_{vi}^t = q_{vi} \oplus \text{FFN} \Big(softmax\Big( \frac{q_{vi}k_d^T}{\sqrt{d_k}}\Big)v_d\Big), \label{eq:CmDRM}
\end{equation}
where $\sqrt{d_k}$ is a scaling factor and $\text{FFN}(\cdot)$ denotes the feed-forward network. Furthermore, given the distinct contributions of primary features and differentially reinforced features to scene characterization, we devise a learnable contribution measurement to integrate both components while adaptively balancing their contributions. As shown in Fig.~\ref{fig:Framework}~(b), $\mathcal{F}_{x}^t$ and $\tilde{\mathcal{F}}_{x}^t$ are first concatenated along the channel dimension. The result is then processed through convolutional layers followed by average pooling to derive contribution metric scores~($w, \tilde{w}$). The weighted features are further refined through channel and spatial attention. The final cross-modal reinforced features are formulated as:
\begin{equation}
	\begin{aligned}		
		\tilde{\mathcal{F}}_{x_{c}}^t & = (\tilde{w} * \tilde{\mathcal{F}}_{x}^t) * \text{CA}(\tilde{w} * \tilde{\mathcal{F}}_{x}^t), \\ \tilde{\mathcal{F}}_{x_{s}}^t &= \tilde{\mathcal{F}}_{x_{c}}^t * \text{SA}(\tilde{\mathcal{F}}_{x_{c}}^t), \ \hat{\mathcal{F}}_x^t = w * \mathcal{F}_{x}^t  \oplus \tilde{\mathcal{F}}_{x_{s}}^t,  \label{eq:CASA}
	\end{aligned}
\end{equation}
$\text{CA}(\cdot)$ and $\text{SA}(\cdot)$ here denote channel and spatial attention.

\subsubsection{Complete Modality-guided Fusion Module}
Although CmDRM enhances unimodal feature representations via cross-modal differential information, it remains insufficient for comprehensive scene characterization. To address this, we propose a complete modality-guided fusion~(CMGF) module to aggregate complementary contexts, as shown in Fig.~\ref{fig:Framework}~(c). We hypothesize that simply summing features across modalities yields generic comprehensive features but lacks modality-specific expressiveness. Thus, we project comprehensive features $\mathcal{F}_c^t = \hat{\mathcal{F}}_{ir}^t + \hat{\mathcal{F}}_{vi}^t$ as public queries~$q_c$ to distill modality-specific information. The modality-specific features are mapped to corresponding keys~($k_{ir}, k_{vi}$) and values~($v_{ir}, v_{vi}$). The feature aggregation process can be formulated as:
\begin{equation}
	\begin{aligned}		
		\mathcal{F}_f^t = q_c &\oplus \text{FFN} \Big( softmax\Big(\frac{q_{c}k_{ir}^T}{\sqrt{d_k}}\Big)v_{ir}\Big) \\
		&\oplus \text{FFN}\Big(softmax\Big(\frac{q_{c}k_{vi}^T}{\sqrt{d_k}}\Big)v_{vi}\Big).\label{eq:fusion}
	\end{aligned}
\end{equation}

\subsubsection{Bi-temporal Co-attention Module}
Compared to frame-wise image fusion, video fusion better exploit inter-frame dependencies to suppress disturbances and maintain temporal coherence. However, solely relying on Conv3d inadequately unleashes the inherent potential of temporal cues. To this end, we propose a bi-temporal co-attention module~(BiCAM) to establish dense cross-frame interactions via mutual attention, as shown in Fig.~\ref{fig:Framework}~(d). Given the current frame feature~$\mathcal{F}_f^t$, we first establish interactions with neighboring frames through a standard multi-head cross-attention mechanism:
\begin{equation}
	\mathcal{A}^{t-1} \!=\! softmax\Big(\!\frac{q^t{k^{t-1}}^T}{\sqrt{d_k}}\!\Big),
	\mathcal{A}^{t+1} \!=\! softmax\Big(\!\frac{q^t{k^{t+1}}^T}{\sqrt{d_k}}\!\Big), \label{eq:cross-attention}
\end{equation}
where $q^t$ is the shared query derived from the current frame feature $\mathcal{F}^t$. The forward key/value $k^{t-1}$/$v^{t-1}$ and backward key/value $k^{t+1}$/$v^{t+1}$ are projected from the previous frame feature~($\mathcal{F}_f^{t-1}$) and subsequent frame feature~($\mathcal{F}_f^{t+1}$). For boundary frames, we replicate the current frame as the neighbor: 
$\mathcal{F}_f^{t-1}=\mathcal{F}^{t}$ when $t=1$ and $\mathcal{F}_f^{t+1}=\mathcal{F}^{t}$ when $t=T$ 
(thus $(k^{t-1},v^{t-1})=(k^{t},v^{t})$ or $(k^{t+1},v^{t+1})=(k^{t},v^{t})$). A co-attention mechanism is then introduced to enable bidirectional cross-temporal dynamic interactions:
\begin{equation}
	\mathcal{A}_{co} = softmax(\mathcal{A}^{t-1} * {\mathcal{A}^{t+1}}). \label{eq:co-attention}
\end{equation}

Finally, we formulate cross-temporal aggregation as:
\begin{equation}
	\hat{\mathcal{F}}_f^{t} = q^t \oplus \text{FFN}(\mathcal{A}_{co} v^{t-1}) \oplus \text{FFN}(\mathcal{A}_{co} v^{t+1}). \label{eq:BiCAM}
\end{equation}
Note that we deploy $N$ consecutive BiCAMs that \emph{allowing each frame to assimilate complementary cues from adjacent frames and access long-range cross-temporal contexts using neighboring frames as mediators}. This is analogous to \emph{the shifted window mechanism} in Swin Transformer~\cite{Liu2021Swin-Transformer}.

\subsubsection{Feature Enhancement and Modality Unmixing}
The enhancement block employs the efficient transformer from Restormer~\cite{Zamir2022Restormer} as its core operator. The fusion decoder and infrared/visible decoder share a homogeneous structure with this block to restore image representations. Moreover, modality unmixing is performed on fusion features via joint channel-spatial attention mechanisms~\cite{Woo2018CBAM}.

\subsection{Loss Functions}
Adhering to the typical image fusion paradigm~\cite{Ma2022SwinFusion}, our framework first constructs intensity loss~$\mathcal{L}_{int}$, gradient loss~$\mathcal{L}_{grad}$, and color loss~$\mathcal{L}_{color}$ to effectively preserve discriminative information from multi-modal inputs. The definitions of  $\mathcal{L}_{int}$, $\mathcal{L}_{grad}$, $\mathcal{L}_{color}$ are as follows:
\begin{align}
	&\textstyle{\mathcal{L}_{int} = \frac{1}{HW} \sum\nolimits^T_{t=1} \| V_f^t - \max(V_{vi}^t, V_{ir}^t) \|_1,} \label{eq:int} \\
	&\textstyle{\mathcal{L}_{grad} = \frac{1}{HW} \sum\nolimits^T_{t=1}\| \nabla V_f^t - \max(\nabla V_{vi}^t, \nabla V_{ir}^t) \|_1,} \label{eq:grad} \\
	&\textstyle{\mathcal{L}_{color} = \frac{1}{HW} \sum\nolimits^T_{t=1} \| \Phi_{\text{CbCr}}(V_f^t) - \Phi_{\text{CbCr}}(V_{vi}^t) \|_1,} \label{eq:color}
\end{align}
where $\{V_{ir}^{t}, V_{vi}^{t}\}_{t=1}^T$ denotes high-quality infrared-visible videos, $\max(\cdot)$ selects salient targets and textures, $\|\cdot\|_1$ is the $l_1$-norm, $\nabla$ denotes the Sobel operator, and $\Phi_{CbCr}(\cdot)$ converts RGB to CbCr. In particular, we use the Y channel of $V_{f}^t$ and $V_{vi}^t$ in $\mathcal{L}_{int}$ and $\mathcal{L}_{grad}$. Moreover, we introduce the scene fidelity loss~$\mathcal{L}_{sf}$ to fully exploit the potential of modality unmixing and infrared/visible decoder, defined as:
\begin{equation}
	\textstyle{\mathcal{L}_{sf} \!=\! \frac{1}{HW}\! \sum_{x \in \{vi, ir\}} \!\sum^T_{t=1}\! (\! \|\hat{V}_x^t\!-\! V_x^{t} \|_1 \!+\! \|\nabla \hat{V}_x^t\! -\!\nabla V_x^{t} |_1 \!).} \label{eq:sf}
\end{equation}

\begin{figure*}[t]
\centering
\includegraphics[width=0.95\linewidth]{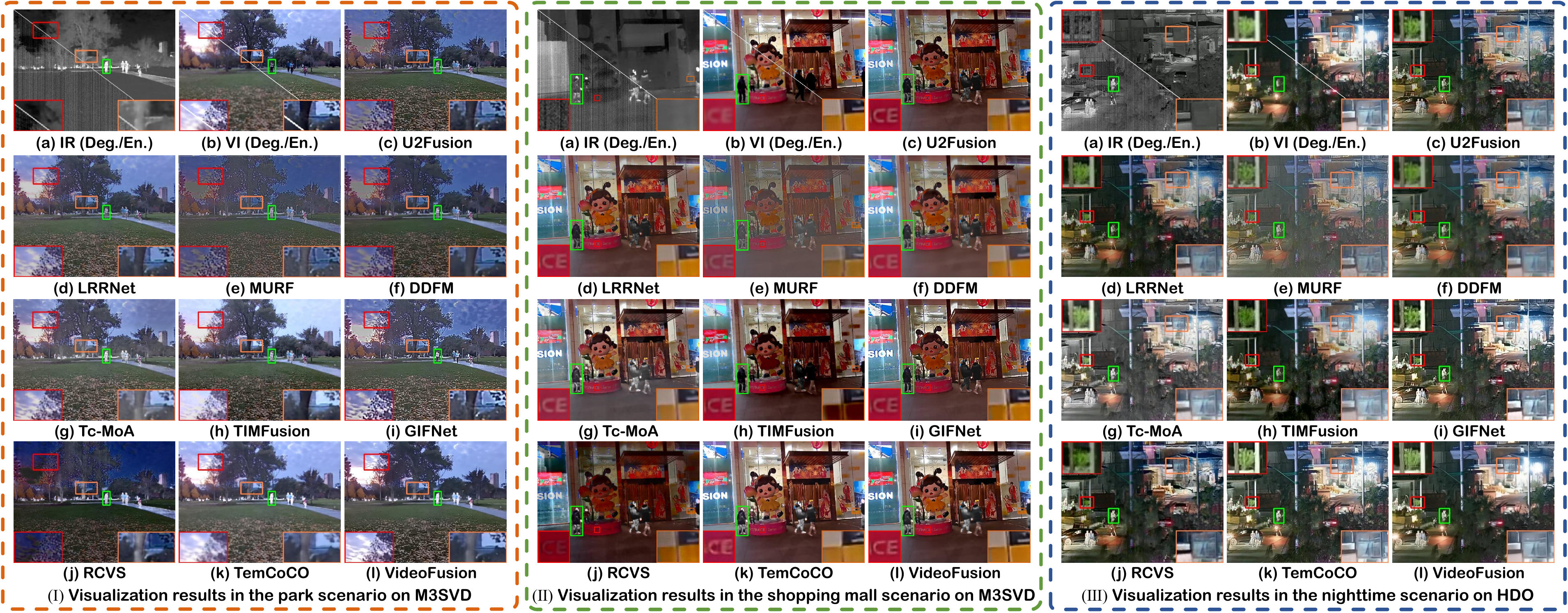}	
\vspace{-0.05in}
\caption{Qualitative comparison results on M3SVD and HDO datasets under degraded scenarios. }
\label{fig:degradation}
\end{figure*}

Besides, we hypothesize that temporal variations of static backgrounds in consecutive frames exhibit local smoothness, while those of dynamic objects in fused or restored videos should align with high-quality sources. This hypothesis can be expressed formally as:
\begin{equation}
	\begin{cases}
		\mathbb{E}\bigl[\Delta V_{x_\mathrm{bg}}^{t}\bigr]  \to 0,  & \text{(static backgrounds)}\\
		\|\Delta V_{f_\mathrm{ob}}^{t} - \Delta V_{x_\mathrm{ob}}^{t}\|_1  \to 0, & \text{(dynamic objects)}
	\end{cases}
	\label{eq:temporal_constraints}
\end{equation}
where $\Delta V_x^t = V_x^{t+1} - V_x^{t}$ and $x \in \{ir, vi\}$. Thus, we design a \textit{variational consistency loss}~$\mathcal{L}_{var}$ to prevent  temporal flickering artifacts in fusion results and restored videos:
\begin{equation}
	\footnotesize
	\textstyle{\mathcal{L}_{var} \!=\! \frac{1}{HW} \!\sum_{x} \!\sum^{T-1}_{t=1} \!(
		\!\underbrace{\|\Delta V_f^t\!-\! \Delta V_x^{t} \|_1}_{\mathcal{L}_{var}^{fu}} \!+
		\!\underbrace{\|\Delta  \hat{V}_x^t \!-\!\Delta V_x^{t} \|_1}_{\mathcal{L}_{var}^{x}}	\!).} \label{eq:var}
\end{equation}

Finally, the total loss of VideoFusion is formulated as:
\begin{equation}
	\begin{aligned}
		\mathcal{L}_{total} = & \lambda_{int}\cdot \mathcal{L}_{int} + \lambda_{grad}\cdot \mathcal{L}_{grad} \ + \\ & \lambda_{color}\cdot \mathcal{L}_{color}+ \lambda_{sf}\cdot \mathcal{L}_{sf} + \lambda_{var}\cdot \mathcal{L}_{var}, \label{eq: total}
	\end{aligned}
\end{equation}
where $\lambda_{int}$, $\lambda_{grad}$, $\lambda_{color}$, $\lambda_{sf}$, and $\lambda_{var}$ are hyper-parameters for balancing various losses.

\begin{table*}[]
	\centering
	\vspace{-0.05in}
	\setlength{\tabcolsep}{3pt}
	\caption{Quantitative comparison on the M3SVD and HDO datasets under degraded scenarios. Each video in M3SVD and HDO contains $200$ and $150$ frames, respectively. The best and second-best results are highlighted in \textcolor{lightred}{\textbf{Red}} and \textcolor[HTML]{7B68EE}{Purple}, respectively.} \label{tab:degradation}
	\resizebox{0.98\textwidth}{!}{
		\begin{tabular}{@{}lclclclclclcccclclcccccccc@{}}
			\toprule
			\multicolumn{1}{l}{} & \multicolumn{12}{c}{\textbf{Multi-Modal   Multi-Scene Video Dataset (M3SVD)}} &  & \multicolumn{12}{c}{\textbf{High-quality   Dual-Optical Dataset (HDO)}} \\ \cmidrule(lr){2-13} \cmidrule(l){15-26} 
			\multicolumn{1}{l}{\multirow{-2}{*}{\textbf{Methods}}} & \textbf{EN~$\uparrow$} & \multicolumn{1}{c}{\textbf{}} & \textbf{MI~$\uparrow$} & \multicolumn{1}{c}{\textbf{}} & \textbf{SD~$\uparrow$} & \multicolumn{1}{c}{\textbf{}} & \textbf{SSIM~$\uparrow$} & \multicolumn{1}{c}{\textbf{}} & \textbf{VIF~$\uparrow$} & \multicolumn{1}{c}{\textbf{}} & \textbf{flowD~$\downarrow$} &  &  & \textbf{EN~$\uparrow$} & \multicolumn{1}{c}{\textbf{}} & \textbf{MI~$\uparrow$} & \multicolumn{1}{c}{\textbf{}} & \textbf{SD~$\uparrow$} & \textbf{} & \textbf{SSIM~$\uparrow$} & \textbf{} & \textbf{VIF~$\uparrow$} & \textbf{} & \textbf{flowD~$\downarrow$} &  \\ \midrule
			\textbf{U2Fusion}~\cite{Xu2022U2Fusion} & 6.904 &  & 2.490 &  & 35.731 & \multicolumn{1}{c}{} & 0.600 & \multicolumn{1}{c}{} & 0.439 & \multicolumn{1}{c}{} & 6.547 &  &  & 7.015 & \multicolumn{1}{c}{} & 1.993 & \multicolumn{1}{c}{} & 48.473 &  & 0.560 &  & 0.385 &  & 8.723 &  \\
			\textbf{LRRNet}~\cite{Li2023LRRNet} & 6.889 &  & 3.120 &  & 38.251 &  & \cellcolor[HTML]{E6E6FA}0.609 &  & 0.452 & \multicolumn{1}{c}{} & 6.874 &  &  & 6.787 & \multicolumn{1}{c}{} & 2.026 & \multicolumn{1}{c}{} & 42.415 &  & 0.523 &  & 0.325 &  & 8.224 &  \\
			\textbf{MURF}~\cite{Xu2023MURF} & 6.185 &  & 1.825 &  & 22.731 &  & 0.543 &  & 0.271 &  & 6.840 &  &  & 6.599 & \multicolumn{1}{c}{} & 1.718 & \multicolumn{1}{c}{} & 32.640 &  & 0.532 &  & 0.324 &  & 9.410 &  \\
			\textbf{DDFM}~\cite{Zhao2023DDFM} & 6.750 &  & 2.656 &  & 32.000 &  & \cellcolor[HTML]{E6E6FA}0.609 &  & 0.448 &  & 6.778 &  &  & 7.019 & \multicolumn{1}{c}{} & 2.017 &  & 42.614 &  & 0.566 &  & 0.381 &  & 7.546 &  \\
			\textbf{TC-MOA}~\cite{Zhu2024TC-MoA} & 7.095 &  & 2.800 &  & 42.412 &  & 0.593 &  & \cellcolor[HTML]{E6E6FA}0.516 &  & 5.102 &  &  & \cellcolor[HTML]{FFC7CE}\textbf{7.431} &  & 2.089 &  & 52.766 &  & 0.543 &  & 0.460 &  & 8.846 &  \\
			\textbf{TIMFusion}~\cite{Liu2024TIMFusion} & 7.063 &  & 3.015 &  & \cellcolor[HTML]{E6E6FA}50.824 &  & 0.580 &  & 0.409 &  & 5.890 &  &  & 6.639 &  & 1.998 &  & 53.146 & \multicolumn{1}{l}{} & 0.481 & \multicolumn{1}{l}{} & 0.339 &  & 7.221 &  \\
			\textbf{GIFNet}~\cite{Cheng2025GIFNet} & 7.042 &  & 2.548 &  & 44.207 &  & 0.580 &  & 0.389 &  & 6.294 &  &  & 7.106 &  & 1.901 &  & \cellcolor[HTML]{FFC7CE}\textbf{59.287} & \multicolumn{1}{l}{} & 0.53 & \multicolumn{1}{l}{} & 0.349 & \multicolumn{1}{l}{} & 9.318 &  \\
			\textbf{RCVS}~\cite{Xie2024RCVS} & 6.551 &  & 1.979 &  & 37.468 &  & 0.585 &  & 0.471 &  & 6.443 &  &  & 7.229 &  & 1.992 &  & 56.277 & \multicolumn{1}{l}{} & 0.519 & \multicolumn{1}{l}{} & 0.415 & \multicolumn{1}{l}{} & 9.677 &  \\
			\textbf{TemCoCo}~\cite{Gong2025TemCoCo} & \cellcolor[HTML]{FFC7CE}\textbf{7.174} &  & \cellcolor[HTML]{E6E6FA}3.548 &  & 50.421 &  & 0.597 &  & 0.490 &  & \cellcolor[HTML]{E6E6FA}4.378 &  &  & \cellcolor[HTML]{E6E6FA}7.352 &  & \cellcolor[HTML]{E6E6FA}2.402 &  & 57.001 & \multicolumn{1}{l}{} & \cellcolor[HTML]{E6E6FA}0.573 & \multicolumn{1}{l}{} & \cellcolor[HTML]{E6E6FA}0.473 & \multicolumn{1}{l}{} & \cellcolor[HTML]{E6E6FA}6.964 &  \\
			\textbf{VideoFusion} & \cellcolor[HTML]{E6E6FA}7.167 &  & \cellcolor[HTML]{FFC7CE}\textbf{4.008} &  & \cellcolor[HTML]{FFC7CE}\textbf{52.465} &  & \cellcolor[HTML]{FFC7CE}\textbf{0.632} &  & \cellcolor[HTML]{FFC7CE}\textbf{0.526} &  & \cellcolor[HTML]{FFC7CE}\textbf{3.294} &  &  & 7.288 &  & \cellcolor[HTML]{FFC7CE}\textbf{2.682} &  & \cellcolor[HTML]{E6E6FA}57.024 & \multicolumn{1}{l}{} & \cellcolor[HTML]{FFC7CE}\textbf{0.582} & \multicolumn{1}{l}{} & \cellcolor[HTML]{FFC7CE}\textbf{0.475} & \multicolumn{1}{l}{} & \cellcolor[HTML]{FFC7CE}\textbf{6.098} &  \\ \bottomrule
		\end{tabular}
	}
\end{table*}

\begin{table}[t]
	\centering
	\vspace{-0.05in}
	\setlength{\tabcolsep}{2pt}
	\caption{Quantitative comparison results on the M3SVD dataset under normal scenarios, where each video contains 200 frames.} \label{tab:normal}	
	\resizebox{0.49\textwidth}{!}{
		\begin{tabular}{@{}lcccccccccccc@{}}
			\toprule
			\textbf{Methods} & \textbf{EN~$\uparrow$} & \textbf{} & \textbf{MI~$\uparrow$} &  & \textbf{SD~$\uparrow$} &  & \textbf{SSIM~$\uparrow$} &  & \textbf{VIF~$\uparrow$} &  & \textbf{flowD~$\downarrow$} &  \\ \midrule
			\textbf{U2Fusion}~\cite{Xu2022U2Fusion} & 6.935 &  & 2.659 &  & 36.316 &  & 0.614 &  & 0.505 &  & 6.615 &  \\
			\textbf{LRRNet}~\cite{Li2023LRRNet} & 6.876 &  & 3.219 &  & 37.736 &  & \cellcolor[HTML]{E6E6FA}0.618 &  & 0.488 &  & 7.473 &  \\
			\textbf{MURF}~\cite{Xu2023MURF} & 6.209 &  & 1.915 &  & 22.844 &  & 0.546 &  & 0.310 &  & 7.373 &  \\
			\textbf{DDFM}~\cite{Zhao2023DDFM} & 6.777 &  & 2.642 &  & 31.960 &  & 0.593 &  & 0.462 &  & 7.064 &  \\
			\textbf{TC-MOA}~\cite{Zhu2024TC-MoA} & 7.109 &  & 2.894 &  & 42.711 &  & 0.602 &  & 0.577 &  & 5.305 &  \\
			\textbf{TIMFusion}~\cite{Liu2024TIMFusion} & 7.085 &  & 3.083 &  & \cellcolor[HTML]{E6E6FA}51.637 &  & 0.590 &  & 0.433 &  & 6.295 &  \\
			\textbf{GIFNet}~\cite{Cheng2025GIFNet} & 6.881 &  & 2.949 &  & 43.829 &  & \cellcolor[HTML]{FFC7CE}\textbf{0.646} &  & 0.495 &  & \cellcolor[HTML]{E6E6FA}4.082 &  \\
			\textbf{RCVS}~\cite{Xie2024RCVS} & 6.592 &  & 2.123 &  & 37.638 &  & 0.606 &  & \cellcolor[HTML]{E6E6FA}0.595 &  & 5.547 &  \\
			\textbf{TemCoCo}~\cite{Gong2025TemCoCo} & \cellcolor[HTML]{E6E6FA}7.174 &  & \cellcolor[HTML]{E6E6FA}3.548 &  & 50.423 &  & 0.597 &  & 0.490 &  & 4.379 &  \\
			\textbf{VideoFusion} & \cellcolor[HTML]{FFC7CE}\textbf{7.199} &  & \cellcolor[HTML]{FFC7CE}\textbf{4.191} &  & \cellcolor[HTML]{FFC7CE}\textbf{53.375} &  & \cellcolor[HTML]{FFC7CE}\textbf{0.646} &  & \cellcolor[HTML]{FFC7CE}\textbf{0.605} &  & \cellcolor[HTML]{FFC7CE}\textbf{3.494} &  \\ \bottomrule
		\end{tabular}
	}
\end{table}

\section{Experiments}
\subsection{Configurations and Implementation Details}
We set $T=7$ during training due to GPU memory limitations and $T=25$ for testing. Other key parameters are set as $[N_1, N_2, N_3] = [2, 2, 4]$ and $[C_1, C_2, C_3] = [32, 64, 128]$. The hyper-parameters balancing various losses are empirically set as $\lambda_{grad}=1$, $\lambda_{sf}=10$,  $\lambda_{int}=15$, $\lambda_{color}=100$, $\lambda_{var}=100$. We train our VideoFusion for $20$ epochs using the AdamW optimizer with $\beta_1=0.9$ and $\beta_2=0.999$, and an initial learning rate of $1\times10^{-4}$ decayed to $1\times10^{-5}$ via cosine annealing. Training is conducted on our M3SVD dataset with $200$ videos, where visible videos are degraded by Gaussian blur (kernel size $15$, standard deviation in $[0.9, 2.1]$) and infrared videos by stripe noise generated via a physics-inspired model~\cite{Cai2024MDIVDNet}. Our method is implemented on PyTorch. All experiments are conducted on NVIDIA RTX~4090 GPUs and a 2.50~GHz Intel Xeon Platinum 8180 CPU.

We compare fusion performance with SOTA image fusion methods, including U2Fusion~\cite{Xu2022U2Fusion}, LRRNet~\cite{Li2023LRRNet}, MURF~\cite{Xu2023MURF}, DDFM~\cite{Zhao2023DDFM}, TC-MoA~\cite{Zhu2024TC-MoA}, TIMFusion~\cite{Liu2024TIMFusion}, and GIFNet~\cite{Cheng2025GIFNet}, as well as video fusion approaches, \emph{i.e.} RCVS~\cite{Xie2024RCVS} and TemCoCo~\cite{Gong2025TemCoCo}, under degraded and normal scenarios. We also employ SOTA video deblurring~(\emph{i.e.}, DSTNet~\cite{Pan2023DSTNet}) and video denoising~(\emph{i.e.}, MDIVDNet~\cite{Cai2024MDIVDNet}) algorithms to pre-enhance degraded videos for a fair comparison. Both MDIVDNet and DSTNet are retrained on our dataset with their default configurations. Six metrics are employed to comprehensively evaluate fusion methods from two perspectives, \emph{i.e.,} fusion quality and temporal consistency. Particularly, EN~\cite{Roberts2008EN}, MI~\cite{Qu2002MI}, SD~\cite{Rao1997SD}, SSIM~\cite{Wang2004SSIM}, and VIF~\cite{Han2013VIF} measure fusion quality, while flowD~\cite{Gong2025TemCoCo} assesses temporal coherence between frames.

\subsection{Fusion Performance Comparison}
Qualitative results under degraded scenarios on M3SVD and HDO datasets are presented in Fig.~\ref{fig:degradation}. As illustrated in Fig.~\ref{fig:degradation}~(\Rmnum{1})-(\Rmnum{2}), U2Fusion, LRRNet, MURF, DDFM, TIMFusion, and GIFNet struggle to preserve prominent targets in the infrared videos, with MURF and TIMFusion introducing artifacts. Among video-based schemes, RCVS introduces more artifacts and reduces the overall brightness distribution of fusion results. TemCoCo, on the other hand, suffers from significant blur, despite it attempts to leverage temporal cues to mitigate degradation. This suggests that its information aggregation and restoration modules still have inherent limitations. In contrast, our VideoFusion effectively integrates complementary information and temporal contexts, producing clearer and more comprehensive scene representations. In nighttime scenes from the HDO dataset, as shown in Fig.~\ref{fig:degradation}~(\Rmnum{3}), video fusion approaches including TemCoCo and VideoFusion not only enhance details but also highlight significant targets, such as pedestrians.

Quantitative evaluations on $20$ videos each from M3SVD and HDO dataset are shown in Tab.~\ref{tab:degradation}. VideoFusion achieves the highest MI and SSIM, indicating that our method effectively transfers complementary information from source videos to fusion results. Meanwhile, the superior VIF demonstrates that VideoFusion better aligns with human visual perception. Comparable EN and SD suggest that our results contain abundant information and high contrast. Among video-based alternative, TemCoCo achieves comparable performance, whereas the frame-by-frame approach RCVS exhibits limited performance on these metrics, further underscoring the necessity of leveraging temporal sequence information. Additional results on normal M3SVD scenarios are reported in Tab.~\ref{tab:normal}, where VideoFusion demonstrates superior performance across all metrics. This further validates its ability to fully exploit cross-modal complementary cues and temporal context to counteract interference and enhance scene characterization.


\begin{figure}[t]
	\centering
	\includegraphics[width=0.98\linewidth]{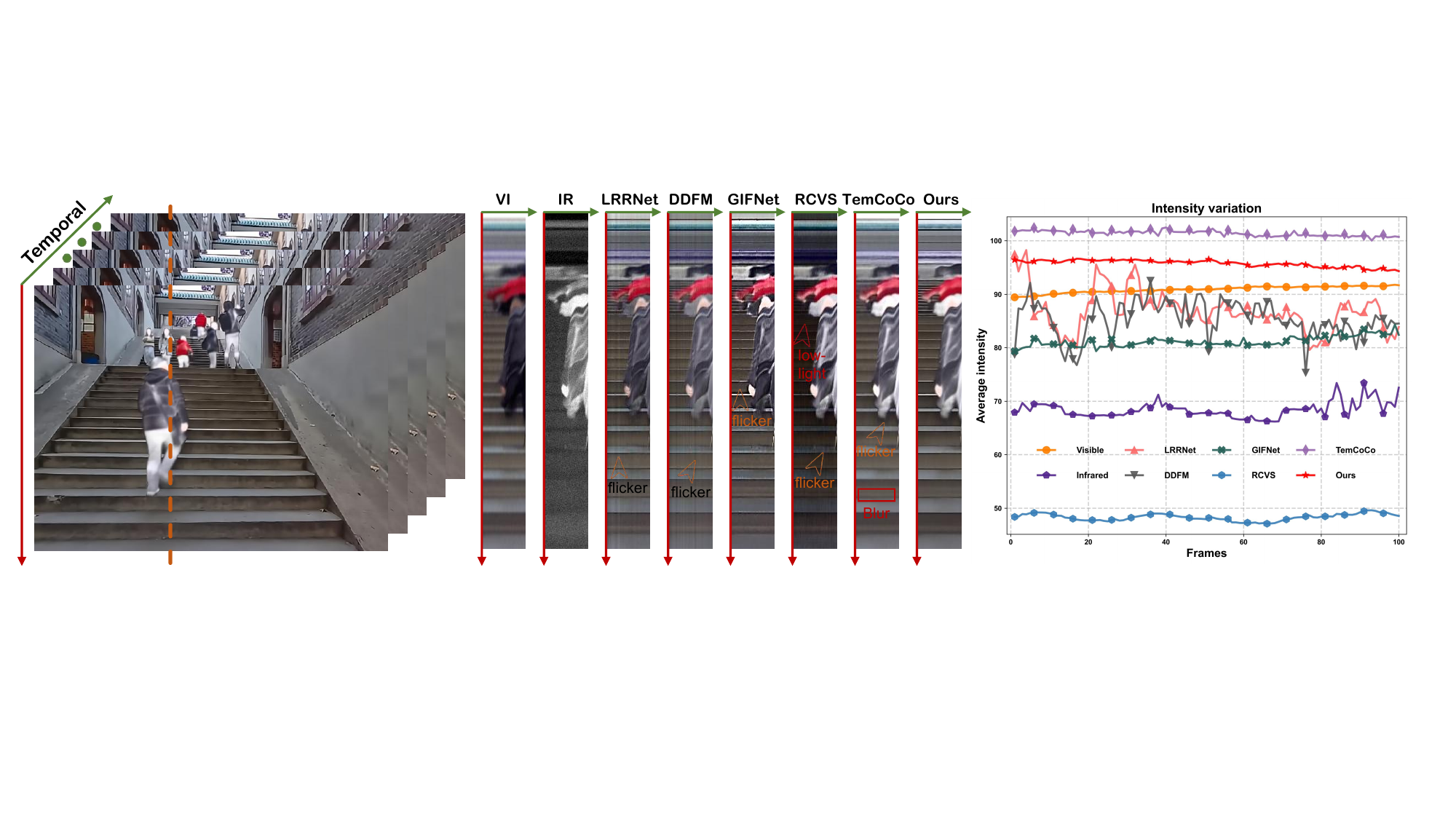}	
	\caption{Temporal consistency comparison of source and fused videos. Following~\cite{Pan2023DSTNet}, column pixels (dotted line) are visualized and their average brightness variation is computed.}
	\label{fig:Temporal}
\end{figure}

\begin{figure}[t]
	\centering
	\includegraphics[width=0.99\linewidth]{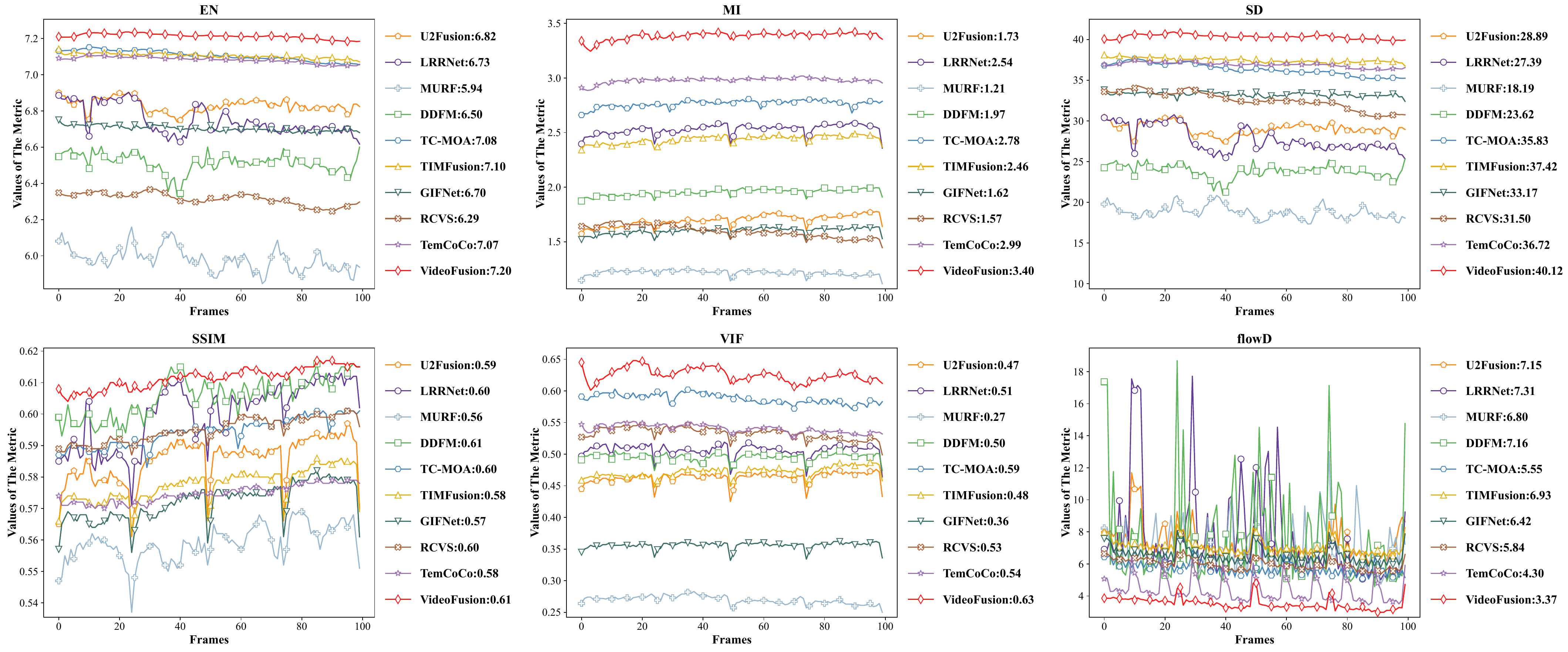}
	\caption{Temporal variation of metrics on sequences.}
	\label{fig:0112_1707}
\end{figure}

\subsection{Temporal Consistency Comparison}
As shown in Tabs.~\ref{tab:degradation} and~\ref{tab:normal}, VideoFusion achieves the lowest flowD on both M3SVD and HDO datasets, indicating its superior temporal stability and consistency. TemCoCo also attains relatively low flowD, further highlighting the significance of temporal modeling. In contrast, frame-by-frame fusion schemes exhibit higher flowD, implying noticeable temporal jitter in their fused videos. Furthermore, following~\cite{Pan2023DSTNet}, we visualize the temporal variations of different videos in Fig.~\ref{fig:Temporal}, where frame-by-frame fusion methods exhibit evident flickering, especially in DDFM and LRRNet. In contrast, our method achieves temporally stable fusion, benefiting from BiCAM and variational consistency loss. The stable mean intensity variation further confirms the superior temporal consistency of our method. Additionally, Fig.~\ref{fig:0112_1707} illustrates the temporal trajectories of various metrics across consecutive sequences, reinforcing this advantage.

\begin{table}[t]
	\centering
	\caption{Comparison of computational efficiency. $^*$ indicates methods incurring extra cost from preprocessing algorithms~(\emph{e.g.}, MDIVDNet and DSTNet) under degraded scenarios.} \label{tab:efficiency}
	\resizebox{0.48\textwidth}{!}{
		\begin{tabular}{@{}l
				>{\columncolor[HTML]{EFEFEF}}c ccccc@{}}
			\toprule
			\textbf{Methods} & \textbf{MDIVDNet} & \textbf{U2Fusion$^*$} & \textbf{LRRNet$^*$} & \textbf{MURF$^*$} & \textbf{DDFM$^*$} & \textbf{TC-MoA$^*$} \\ \midrule
			\textbf{Parm. (M)} & 10.047 & 0.659 & 0.049 & 0.116 & 552.660 & 340.354 \\
			\textbf{Flops (G)} & 240.15 & 405.2 & 14.17 & 31.49 & 5220.5 & 3932.09 \\
			\textbf{Time (s)} & 0.033 & 0.077 & 0.011 & 0.195 & 34.502 & 0.183 \\ \midrule
			\textbf{Methods} & \textbf{DSTNet} & \textbf{TIMFusion$^*$} & \textbf{GIFNet$^*$} & \textbf{RCVS$^*$} & \textbf{TemCoCo} & \textbf{VideoFusion} \\ \midrule
			\textbf{Parm. (M)} & 7.448 & 1.232 & 3.291 & 0.670 & 19.183 & 6.743 \\
			\textbf{Flops (G)} & 224.38 & 215.9 & 186.59 & 27.88 & 491.14 & 267.78 \\
			\textbf{Time (s)} & 0.021 & 0.029 & 0.043 & 0.012 & 0.052 & 0.067 \\ \bottomrule
		\end{tabular}
	}
\end{table}

\begin{figure}[t]
	\centering
	\includegraphics[width=0.98\linewidth]{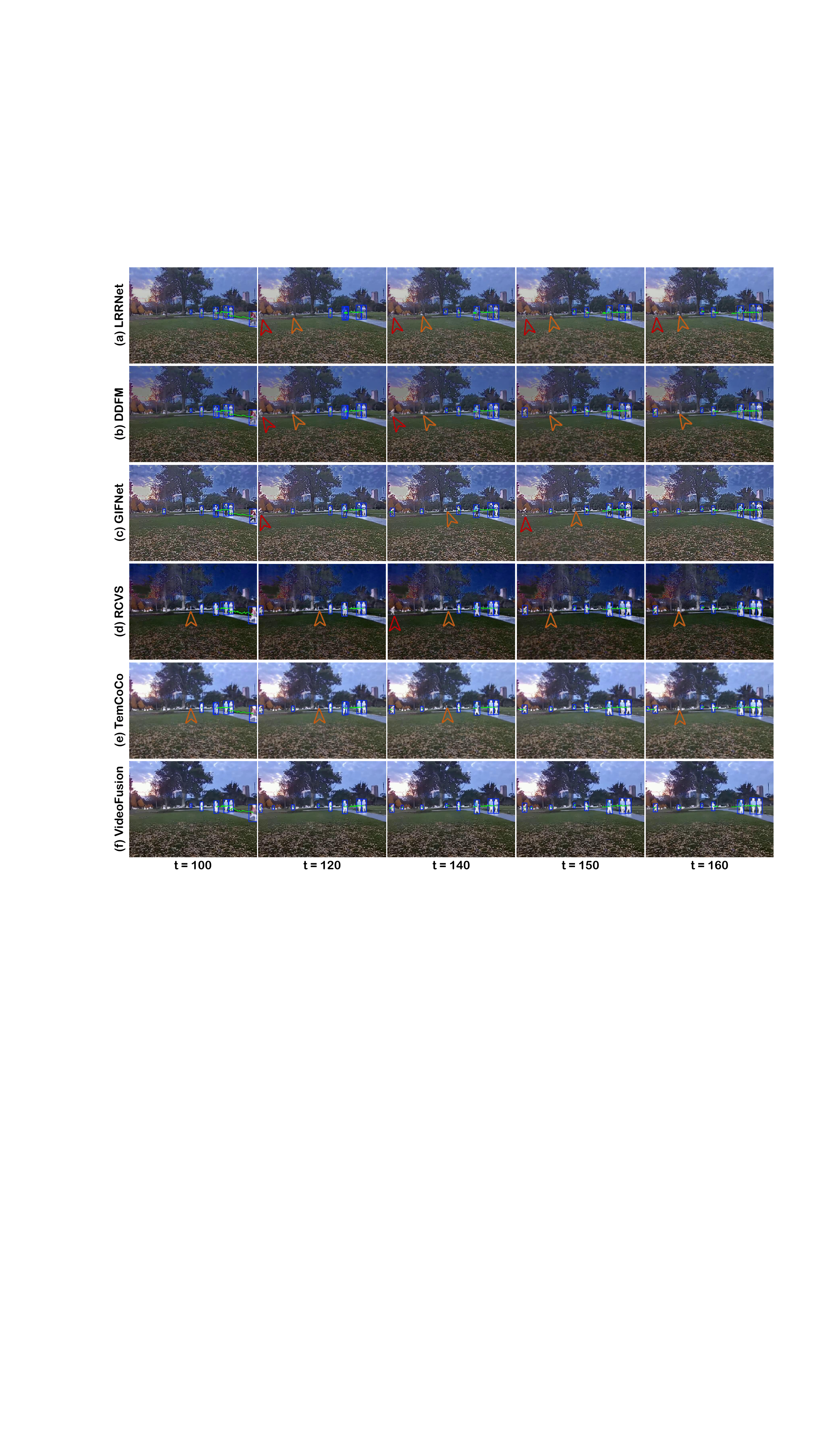}	
	\caption{Visual comparison of object track on M3SVD dataset.}
	\label{fig:tracking}
\end{figure}


\subsection{Extension Experiments}
\textbf{Object Tracking.} Effective information enhancement and aggregation not only enhance visual perception, but also improve machine vision. Fig.~\ref{fig:tracking} provides a qualitative assessment of various fusion schemes for object tracking on M3SVD, where  a pre-trained YOLO v11~\cite{Khanam2024Yolov11} serves as the baseline tracker. On the one hand, the tracker detects more objects in our results, as our method effectively integrates cross-modal complementary information and temporal context, providing a more comprehensive scene representation. On the other hand, it predicts smoother trajectories on our results, benefiting from the improved temporal consistency.

\noindent\textbf{Computational Efficiency.} As shown in Tab.~\ref{tab:efficiency}, VideoFusion achieves computational efficiency comparable to image-level fusion methods. Notably, in degraded scenarios, most algorithms incur additional computational overhead due to preprocessing steps, further demonstrating the efficiency advantage of our VideoFusion.

\begin{figure}[t]
	\centering
	\includegraphics[width=0.99\linewidth]{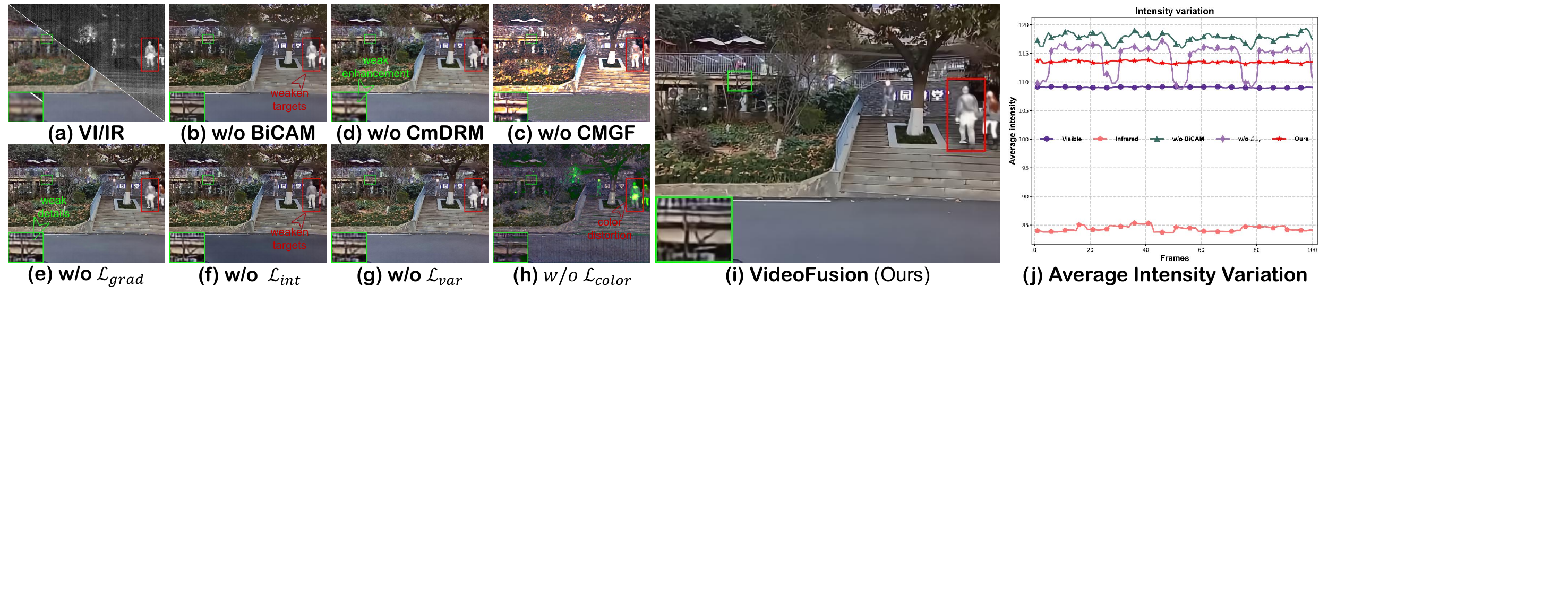}	
	\caption{Visual comparison of ablation studies.}
	\label{fig:Ablation}	
	\vspace{-0.05in}
\end{figure}

\begin{figure}[t]
	\centering
	\includegraphics[width=0.99\linewidth]{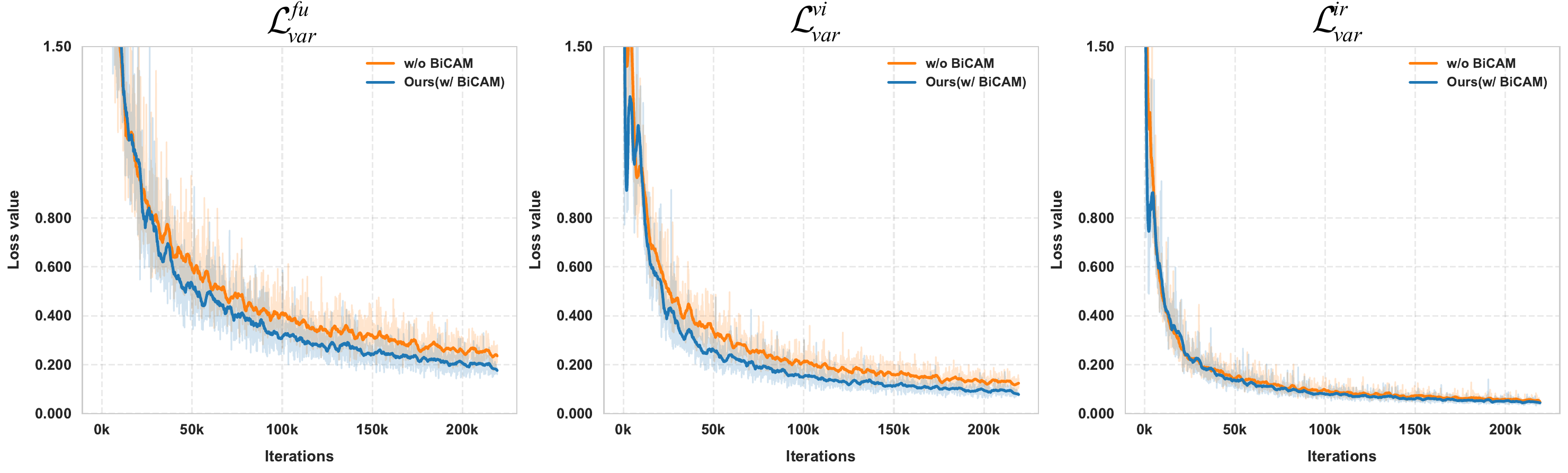}	
	\caption{Training processes for BiCAM.}
	\label{fig:Loss}
\end{figure}

\noindent\textbf{Ablation Studies.}
As shown in Fig.~\ref{fig:Ablation}, we conduct a series of ablation studies to validate our key designs. Removing BiCAM or $\mathcal{L}_{int}$ weakens salient targets, while omitting CmDRM reduces information recovery. Besides, replacing CMGF with a simple summation causes severe distortion. Similarly, w/o $\mathcal{L}_{color}$ induces noticeable color distortion and artifacts, and w/o $\mathcal{L}_{grad}$ results in detailed texture loss. Moreover, as shown in Fig.~\ref{fig:Ablation}~(j), removing BiCAM or $\mathcal{L}_{var}$ disrupts temporal consistency. Particularly, as presented in Fig.~\ref{fig:Loss}, w/o BiCAM affects the convergence of $\mathcal{L}_{var}$. Although Tab.~\ref{tab:ablation} shows that the model without CMGF achieves optimal SD, this comes at the cost of introducing significant artifacts and distortions. In contrast, VideoFusion with these key components achieves more balanced overall performance while preserving temporal coherence.

\begin{table}[t]
	\centering
	\caption{Quantitative comparison of ablation studies.} \label{tab:ablation}	
	\setlength{\tabcolsep}{2pt}
	\resizebox{0.47\textwidth}{!}{
		\begin{tabular}{@{}lcccccccccccc@{}}
			\toprule
			\textbf{Configs} & \multicolumn{1}{c}{\textbf{EN~$\uparrow$}} &  & \multicolumn{1}{c}{\textbf{MI~$\uparrow$}} &  & \multicolumn{1}{c}{\textbf{SD~$\uparrow$}} &  & \multicolumn{1}{c}{\textbf{SSIM~$\uparrow$}} &  & \multicolumn{1}{c}{\textbf{VIF~$\uparrow$}} &  & \multicolumn{1}{c}{\textbf{flowD~$\downarrow$}} &  \\ \midrule
			\textbf{w/o BiCAM} & \cellcolor[HTML]{FFC7CE}\textbf{7.250} &  & 3.439 &  & 52.194 &  & 0.601 &  & 0.472 &  & 4.747 &  \\
			\textbf{w/o CmDRM} & 6.953 &  & 3.557 &  & 48.439 &  & 0.612 &  & \cellcolor[HTML]{E6E6FA}0.510 &  & 3.728 &  \\
			\textbf{w/o CMGF} & 7.046 &  & 2.099 &  & \cellcolor[HTML]{FFC7CE}\textbf{61.403} &  & 0.366 &  & 0.233 &  & 7.029 &  \\
			\textbf{w/o $\mathcal{L}_{grad}$} & 7.208 &  & \cellcolor[HTML]{E6E6FA}3.702 &  & 52.239 &  & 0.615 &  & 0.5 &  & \cellcolor[HTML]{E6E6FA}3.669 &  \\
			\textbf{w/o $\mathcal{L}_{int}$} & 7.106 &  & 2.985 &  & 47.630 &  & \cellcolor[HTML]{E6E6FA}0.630 &  & 0.468 &  & 3.684 &  \\
			\textbf{w/o $\mathcal{L}_{var}$} & \cellcolor[HTML]{E6E6FA}7.211 &  & 3.432 &  & 52.245 &  & 0.599 &  & 0.480 &  & 6.056 &  \\
			\textbf{w/o $\mathcal{L}_{color}$} & 6.839 &  & 2.102 &  & 39.854 &  & 0.457 &  & 0.240 &  & 6.031 &  \\
			\textbf{VideoFusion} & 7.167 &  & \cellcolor[HTML]{FFC7CE}\textbf{4.008} &  & \cellcolor[HTML]{E6E6FA}52.465 &  & \cellcolor[HTML]{FFC7CE}\textbf{0.632} &  & \cellcolor[HTML]{FFC7CE}\textbf{0.526} &  & \cellcolor[HTML]{FFC7CE}\textbf{3.294} &  \\ \bottomrule
		\end{tabular}
	}
\end{table}

\section{Conclusion}
This work has presented \textbf{M3SVD}, a multi-modal multi-scene video dataset, and \textbf{VideoFusion}, a spatio-temporal collaborative framework for multi-modal video fusion. On the one hand, a cross-modal differential reinforcement module has been devised for cross-modal information interaction and enhancement, while a complete modality-guided fusion strategy has been employed to integrate multi-modal features. On the other hand, a bi-temporal co-attention mechanism with a variational consistency loss has been designed to dynamically aggregate forward-backward temporal contexts, reinforcing cross-frame feature representations. Extensive experiments have demonstrated the superiority of VideoFusion over conventional image-based fusion paradigms in sequential scenarios, particularly in alleviating temporal inconsistency and interference.

{
    \small
    \bibliographystyle{ieeenat_fullname}
    \bibliography{aaai2026}
}


\end{document}